\documentclass[lettersize,journal]{IEEEtran}
\usepackage{enumerate}
\usepackage[pdftex]{graphicx}
\usepackage{amsmath}

\usepackage{multirow}
\usepackage{cite}
\usepackage{amsfonts}
\usepackage{tcolorbox}
\usepackage[normalem]{ulem}
\usepackage{amssymb}

\usepackage{graphicx}
\usepackage{subfigure}
\usepackage{caption}
\usepackage{url}
\usepackage{array}
\newcommand{\PreserveBackslash}[1]{\let\temp=\\#1\let\\=\temp}
\newcolumntype{C}[1]{>{\PreserveBackslash\centering}p{#1}}
\newcolumntype{R}[1]{>{\PreserveBackslash\raggedleft}p{#1}}
\newcolumntype{L}[1]{>{\PreserveBackslash\raggedright}p{#1}}
\usepackage{multirow}
\usepackage{booktabs}
\usepackage{setspace}
\usepackage{algorithm}
\usepackage{algorithmicx}
\usepackage{algpseudocode}
\usepackage{gensymb}
\usepackage{balance}
\usepackage{tabularx}
\usepackage{ragged2e}
\usepackage{cite}
\usepackage{pifont}

\begin{document}

\title{LiDAR-based HD Map Localization using Semantic Generalized ICP with Road Marking Detection}
% \title{Localization on an HD Map with Road Marking Detection and SG-ICP Registration with LiDARs}

\author{Yansong Gong, Xinglian Zhang, Jingyi Feng, Xiao He and Dan Zhang$^*$
        % <-this % stops a space
\thanks{Yansong Gong (yansong.gong@uisee.com), Xinglian Zhang, Jingyi Feng, Xiao He and Dan Zhang (corresponding author, dan.zhang@uisee.com) are with UISEE Technology (Beijing) Co., Ltd.}
\thanks{This version of the manuscript has been accepted by IEEE/RSJ International Conference on Intelligent Robots and Systems (\textbf{IROS 2024}).}

}



\maketitle

\begin{abstract}
% Accurate vehicle localization is vital for autonomous driving.
In GPS-denied scenarios, a robust environmental perception and localization system becomes crucial for autonomous driving.
%LiDAR sensors offer precise 3D representation irrespective of varying light conditions.
%Road markings, identified from LiDAR point clouds due to their high reflectance, present a challenge in real-time urban environmental perception.
%Low point cloud density necessitates aggregating multiple scans for robust road marking detection.
%However, direct aggregation poses challenges in meeting real-time requirements.
%Balancing the need for denser data with real-time performance is a critical consideration.
In this paper, a LiDAR-based online localization system is developed, incorporating road marking detection and registration on a high-definition (HD) map.
Within our system, a road marking detection approach is proposed with real-time performance, in which an adaptive segmentation technique is first introduced to isolate high-reflectance points correlated with road markings, enhancing real-time efficiency.
Then, a spatio-temporal probabilistic local map is formed by aggregating historical LiDAR scans, providing a dense point cloud.
% Notably, higher probability values are assigned to newly observed points, mitigating accumulated errors.
Finally, a LiDAR bird's-eye view (LiBEV) image is generated, and an instance segmentation network is applied to accurately label the road markings.
For road marking registration, a semantic generalized iterative closest point (SG-ICP) algorithm is designed.
Linear road markings are modeled as 1-manifolds embedded in 2D space, mitigating the influence of constraints along the linear direction, addressing the under-constrained problem and achieving a higher localization accuracy on HD maps than ICP.
Extensive experiments are conducted in real-world scenarios, demonstrating the effectiveness and robustness of our system.

\end{abstract}

\begin{IEEEkeywords}
Localization, autonomous vehicles, road markings, LiDAR, HD map.
\end{IEEEkeywords}

\section{Introduction}
\IEEEPARstart{A}{ccurate} localization is a prerequisite for autonomous driving.
%In most cases, a GPS can provide a centermeter-level pose of the vehicle using the real-time kinematic (RTK) technique.
In unsheltered open-air environments, the global positioning system (GPS) is the predominant technology for accurate localization.
%However, poses provided by the GPS are not stable when signals from satellites are occluded by ceilings or viaducts.
However, the GPS-provided poses become unstable when satellite signals are obstructed by ceilings or viaducts.
Therefore, localization through environmental perception using observation sensors, such as cameras and light detection and ranging (LiDAR) sensors, becomes necessary for autonomous vehicles, especially in GPS-denied environments.
% Currently, HD maps have been developed in the field of autonomous driving, containing rich semantic information of environments.

%localization through environmental perception using observation sensors (e.g. cameras and LiDARs) is necessary for autonomous vehicles, especially in GPS-denied environments.
% Therefore, in GPS-denied scenarios, a fast and robust environmental perception system is fundamental to accurate localization for autonomous vehicles.

In autonomous vehicle navigation, the detection of road markings stands out as the most widely employed technique for achieving precise and stable environmental perception. Subsequently, the detected road markings can be associated with semantic elements in high-definition (HD) maps to estimate the vehicle's pose.
Cameras have been widely used for road marking detection \cite{ITSC2020lane,ITSC2020detecting,TVT2020robust,TITS2023lane}, because camera images contain rich texture information of environments. However, cameras are limited by the susceptibility of illumination variations and distortions in bird's-eye view (BEV) lane representation, rendering them less robust for certain applications. \cite{IROS2018deep,Wang2014AutomaticPB}.
% Additionally, converting camera images into a 2D bird's eye view (BEV) may introduce distortions in the representation of lane lines, potentially leading to challenges in driving tasks like path planning \cite{IROS2018deep,Wang2014AutomaticPB}.

\begin{figure}[htbp]
	\centering
	\includegraphics[width=1.0\linewidth]{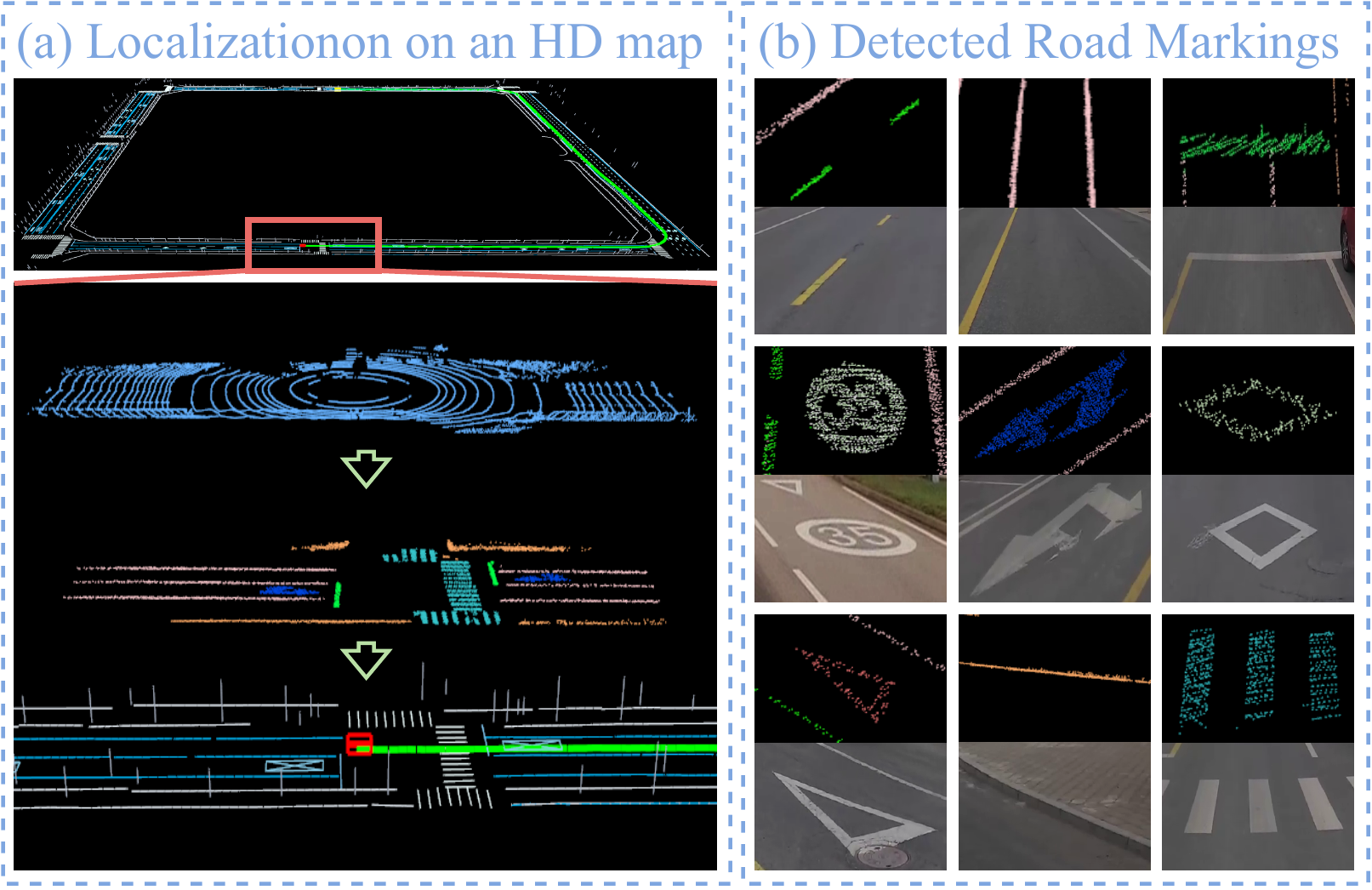}
	\caption{
	(a) The HD map localization of our approach is visualized, where the trajectory of vehicle localization is marked in green, and the current pose of the vehicle is represented by a red cube. The blue point cloud represents ground points from a single-frame LiDAR data. These ground points are adaptively segmented to identify highly reflective points. Subsequently, they are aggregated by successive frames of data to form a denser point cloud. Finally, semantic segmentation is applied to obtain a semantic point cloud, which is then registered with the HD map to estimate the vehicle's pose.
	(b) Road markings extracted using our approach are visualized, encompassing dashed lanes, solid lanes, stop lines, texts, arrows, diamond signs, triangle signs, curbs, and crosswalks.
	}
	\label{marking_vis}
\end{figure}

In contrast, LiDAR sensors exhibit reduced sensitivity to varying illumination conditions and provide a precise 3D representation of the environment.
Meanwhile, road markings can be extracted from road surfaces using LiDAR point clouds, leveraging the characteristic of their high reflectance from the retro-reflective materials \cite{RIVEIRO2015,SOILAN2017}.
% LiDAR-based road marking detection problem has been extensively studied in the high-definition (HD) map generation \cite{TITS2022automated,TITS2022road,TITS2022robust,TITS2021capsule,ITSC2016lane,TITS2015using}.
% However, two challenges arise in the context of real-time online urban environmental perception for localization tasks.
% The first challenge lies in the lower point cloud density of individual frame LiDAR data, making it difficult to establish a stable road marking detection method.
% This challenge necessitates the aggregation of multiple scans to create a denser point cloud, enabling more robust detection.
% Second, given the demand for real-time requirement, it is impractical to directly handle a dense point cloud aggregated by multiple scans.
However, these LiDAR-based methods face challenges to balance the need for denser point cloud with the essential requirement for real-time performance.

To address the challenges, a real-time LiDAR-based approach is proposed for road marking detection and registration with HD maps, as visualized in Fig. \ref{marking_vis} (a). For road marking detection, an adaptive segmentation technique is first employed to efficiently isolate points correlated with road markings. Then, a spatio-temporal probabilistic local map is established by aggregating segmented points from historical scans, resulting in a dense point cloud representation of road markings.
Finally, a LiDAR bird's-eye view (LiBEV) image is generated by partitioning the local map into grid cells, and a proficiently trained instance segmentation network (CenterMask\cite{centermask} is selected in our implementation) is applied to accurately detect 9 different types of road markings, as shown in Fig. \ref{marking_vis} (b).

As for the road marking registration,
a semantic generalized iterative closest point (SG-ICP) algorithm is specifically  designed to robustly align the detected road markings with the HD map by leveraging both their semantic and geometric attributes.
% Given the linear nature of certain road markings (such as lane lines, curbs, sidewalk, etc.), an under-constrained problem emerges due to the absence of constraints along the longitudinal.
% This deficiency may lead to inaccuracies in the localization process, potentially yielding erroneous results.
In the proposed SG-ICP registration, the linear types of road markings are modeled as 1-manifolds embedded in the 2D space, making the constraints along the linear direction have minimal influence on the ultimate solution.

%In order to solve above problems, a LiDAR-based localization method with road marking segmentation is proposed, abbreviated as RMS-Loc, which contains four phases, i.e.\ LiDAR bird's-eye view (LiBEV) image generation, road marking segmentation, pose estimation and degradation treatment.
%
%For the road marking segmentation, LiDAR data from adjacent frames are first aggregated through the odometry provided by a speedometer.
%Then a LiBEV image is generated by projecting LiDAR points to the ground plane.
%Next, centermask[] is applied for segmenting road markings on LiBEV images.
%Finally, LiDAR points of ground markings are then extracted according to segmented masks.
%
%For pose estimation, the RMS-Loc first matches the ground plane to the map, solving a 3-DoF pose.
%Then, it utilizes segmentation results of road marking to associate with the corresponding vector in the HD map.
%For different types of markings, cost functions with a unified form are constructed using the point distribution of markings to solve the remained 3-DoF pose.
%In addition, the RMS-Loc gives a constraint analysis of the associated road marking.
%Even if these markings are not enough to constrain the 6-DoF pose, the degradation are eliminated by fusing speedometer data.

The contributions of this paper are summarized as follows.
\begin{enumerate}
  \item
  A LiDAR-based road marking detection approach is proposed for online environmental perception, in which point density and real-time performance are balanced by adaptively segmenting high-reflectance points and updating spatio-temporal probabilistic local map. Finally, a LiBEV image is generated, and 9 different types of road markings can be detected accurately using an instance segmentation network on the LiBEV image.
%   The adaptive segmentation effectively removes a substantial volume of low-reflectance data points associated with the road surface, leading to an informative and dense point cloud, while maintaining real-time performance with tolerable information loss.
%   A dense probabilistic local map is then formed by aggregating the point cloud over historical LiDAR scans. Notably, elevated probability values are assigned to points newly observed, effectively minimizing the impact of accumulated errors over time.
%   Finally, a LiBEV image is generated, and then road marking is detected using a semantic segmentation network.
  \item
%   A novel road marking registration method is proposed for the localization of an autonomous vehicle on an HD map.
%   The registration of road markings presents an under-constrained problem, especially for linear-shaped markings, which may lead to inaccuracies or failures in the localization process.
%   To address this challenge, the SG-ICP algorithm is proposed, representing linear road markings as 1-manifolds embedded in 2D space. This representation allows for solving the registration problem with minimal influence on the under-constrained dimensions, providing a more robust and accurate solution.
  A novel road marking registration algorithm is proposed for localization of autonomous vehicles on HD maps, in which linear road markings are represented as 1-manifolds embedded in 2D space. This representation can provide a robust and accurate solution for the registration problem with minimal influence on the under-constrained dimensions. Compared with the widely-used ICP, SG-ICP achieves higher accuracy of localization.

  \item
  Comprehensive experiments are conducted in real-world scenarios, demonstrating real-time performance and localization accuracy of our system. Furthermore, experimental results indicate the approach's adaptability to various types of LiDAR sensors, as well as its robustness under different vehicle speeds and weather conditions.
%  \item A method is proposed for segmenting road marking with LiDAR-data. This method can not only detect and segment the lane markings on the ground, but also support multi-type road marking segmentation, including stop lines, arrow markings, triangle signs, side walks and slowdown sign.
%  \item The RMS-Loc is proposed which utilizes LiDAR data and HD maps to perform localization on autonomous vehicles. The RMS-Loc segments road marking from LiDAR data and matches them with vectors in the HD map. The 6-DoF pose is efficiently solved by a two-stage pose estimation. Moreover, A unified form of cost function is proposed according to the point distribution of different types of road markings.
%  \item A detailed constraint analysis is given according to the associated road markings. When the localization result is degraded with 1-DoF constraint, the vehicle will not lose its localization by fusing the result provided by speedometer.
\end{enumerate}

\section{Related work}

% In recent years, LiDAR-based environmental perception and localization approaches based on HD maps have been extensively researched.
% %These related works are introduced following according to the type of observation sensors, e.g. cameras and LiDARs.
% In contrast to image-based methods, LiDAR sensors possess a distinct advantage in environmental perception and localization due to less sensitivity to illumination and weather conditions.
% The infrared signals from LiDAR sensors are less susceptible to visible lights, ensuring that adverse lighting conditions have minimal impact on their measurement capabilities.
% Additionally, unlike images, LiDAR point clouds remain free from distortions in the BEV projection.

In urban autonomous driving scenarios, the detection of road markings stands out as a crucial method for environmental perception.
The road markings, typically painted on asphalt roads using retro-reflective materials, play a vital role in guiding autonomous vehicles.
Leveraging the near-infrared wavelength of laser pulses, road markings exhibit higher reflectance compared to unpainted road surfaces \cite{RIVEIRO2015}.
As a result, the LiDAR sensor's ability to capture intensity measurements becomes instrumental in detecting these road markings \cite{SOILAN2017}.
% By exploiting the distinctive reflectance characteristics in the near-infrared spectrum, the LiDAR can effectively detect the presence of road markings, enhancing the vehicle's capacity for the perception and localization.

LiDAR-based road marking detection is extensively applied in the generation of HD maps \cite{TITS2022automated,TITS2022road,TITS2022robust,TITS2021capsule,ITSC2016lane,TITS2015using}.
% Many of these techniques involve the utilization of one or more high-performance LiDAR instruments, typically mounted on the roof of a vehicle and positioned perpendicular to the road level.
Since the data for HD map generation is processed offline, consecutive scans are aggregated into a point cloud with a significantly high density of points, capturing detailed information about the surroundings \cite{TITS2022automated}. However, processing such high-density points is time-consuming, rendering existing methods applied in HD map generation impractical for the online environmental perception and localization.

In existing studies focused on real-time perception, the detection of road markings is achieved by thresholding the measured intensities within a single LiDAR scan.
A lane markings detection approach was developed by Team AnnieWAY for the DARPA Urban Challenge 2007, which detected the painted lane markings from the single scans by thresholding the points with high-reflectance gradients \cite{IV2008lidar}.
Similarly, the approach proposed in \cite{ITSC2009multi} detected highly reflective lane markings by employing a polar lane detector grid.
In \cite{ITSC2014road}, a modified Otsu thresholding technique was employed to segment high-reflectance points obtained from a multilayer LiDAR into distinct categories, such as asphalt and road markings.
Due to the sparsity of LiDAR measurements, the single-scan-based approaches face challenges detecting complete road markings, making the detection results susceptible to noise and lack robustness.

The approach proposed in \cite{ITSC2018lidar} accumulated two consecutive frames of segmented road points, and then applied a fixed intensity threshold to isolate lane marking points.
In the subsequent works \cite{IV2019lidar,ITSC2019lidar}, the approach was extended to detect various types of high-reflectance landmarks, such as road signs and guard rail reflectors, to improve the localization accuracy.
% All the above-mentioned approaches are based on a denser point cloud accumulated by multi frames.
However, these multi-scan-based methods utilize a fixed intensity threshold to segment road marking points, which is sensitive to changes in environmental conditions and sensor types.
% Moreover, these methods often lack generalization across different road types, markings and geographical locations, making it challenging to develop a universally applicable system.

Recently, the deep learning approaches have been widely-used in the road marking detection tasks.
The global feature correlation (LLDN-GFC) was introduced in \cite{CVPR2022klane} which leveraged the spatial characteristics of lane lines within the point cloud including sparsity, thinness, and elongation across the entirety of the ground points.
This method was further improved in \cite{ITSC2022row}, resulting in a substantial reduction in computational cost.
Nevertheless, LLDN-GFC focuses solely on extracting lane lines, overlooking other types of road markings.
This limitation implies that the extracted lane lines can only provide lateral constraints on the vehicle's poses, potentially contributing to a degeneracy problem during the localization.

\section{Methodology}

\begin{figure*}[htbp]
	\centering
	\includegraphics[width=1.0\textwidth]{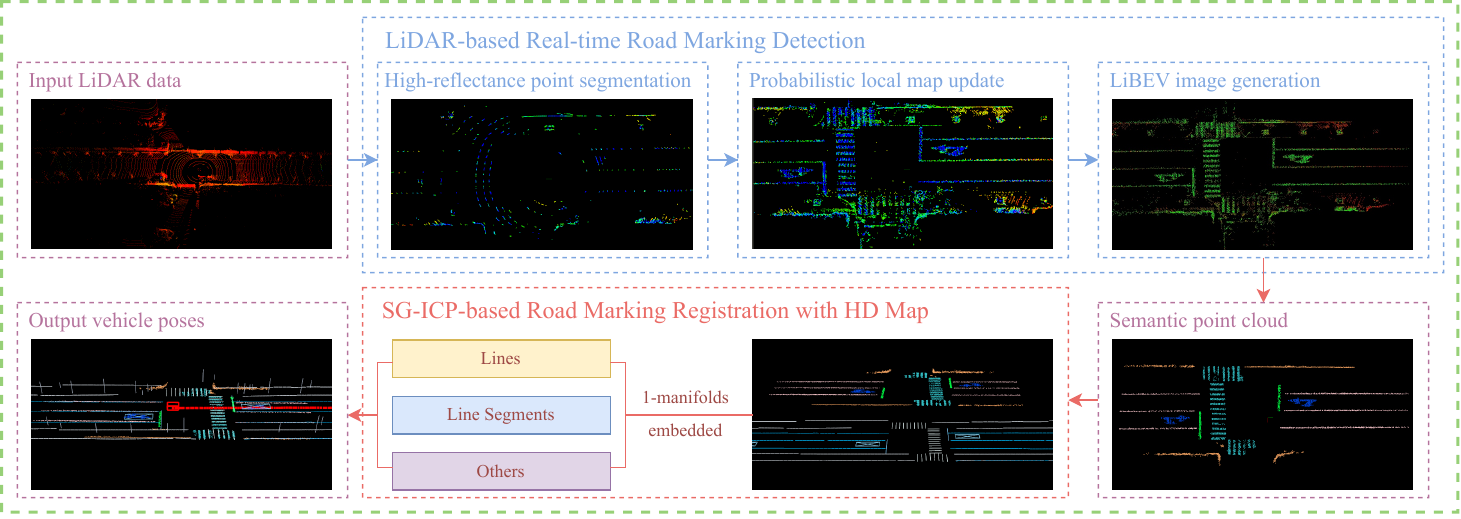}
		\caption{The flowchart of the proposed approach. }
	\label{flowchart}
\end{figure*}

In response to the limitations identified in previous researches, we propose a LiDAR-based road marking detection system for real-time environmental perception.
Additionally, a novel road marking registration algorithm is introduced to enhance the localization accuracy of autonomous vehicles with HD maps.
The flowchart of the proposed system is illustrated in Fig. \ref{flowchart}.

%The LiDAR data undergoes processing through sequential sub-steps, namely high reflectance point segmentation and probabilistic local map update, to generate LiBEV images.
%Subsequently, a highly effective semantic segmentation network is applied to produce semantic mask images on the LiBEV representations.
%Using the generated LiBEV images, a semantic point cloud is derived, incorporating semantic labels for each point.
%In the next phase, the semantic point cloud is associated with corresponding elements in the HD map.
%Three distinct types of elements undergo point cloud registration utilizing SG-ICP.
%The vehicle's pose is then determined by optimizing a cost function constructed through SG-ICP, facilitating accurate alignment and registration of the semantic point cloud with the HD map elements.

\subsection{LiDAR-based Real-Time Road Marking Detection}

Limited by the sparse distribution of LiDAR points, the stable and robust detection of road markings proves challenging when relying solely on individual frame of data.
To overcome this limitation, successive LiDAR scans are aggregated into a local map, generating a denser point cloud that is conducive to effective road marking detection.
In consideration of online requirements and high-reflectance road markings, the aggregation process can selectively extract points with higher intensities from the ground plane.
This approach ensures the construction of a local map optimized for road marking detection, striking a balance between computational efficiency and information richness.

%\subsubsection{Ground Point Segmentation}
%
%Ground points are extracted using a method presented in , in which a part of the ground is treated as a plane. The points with lower $z$-axis values are first extracted as seed points. Then, a plane is fitted using a principal component analysis (PCA). Next, points which are close enough to the fitted plane are extracted as new seed points and used to refit the plane until the plane parameters is stable enough.

\subsubsection{High-Reflectance Point Segmentation}

This procedure aims to adaptively identify points with high reflectance, which are often correlated with road markings painted using retro-reflective materials.
To ensure adaptability across diverse sensors and scenarios, we introduce an adaptive segmentation approach designed to isolate high-reflectance points. This enhancement contributes to a more robust system overall.
% This adaptive methodology ensures a more intelligent segmentation, accommodating variations in sensor characteristics and environmental conditions for improved robustness of the system.

For the efficiency of the system, only ground points are considered, which are extracted from the LiDAR scan utilizing the methodology detailed in \cite{groundsegment}. This approach segments ground points based on height information and subsequently extracts them by partially fitting the ground plane.
Then, a segmentation coefficient $\rho_k$ is introduced to distinguish high-reflectance points from the ground points in the $k$-th scan.
Specifically, points with intensities below $\rho_k$ are excluded from the scan.
Notably, the segmentation coefficient $\rho_k$ is not predetermined manually. Instead, it is dynamically estimated and continuously updated using a Kalman filter.
The state of the Kalman filter is evolved according to the state-transition model
\begin{equation}
	\rho_k=\rho_{k-1}+w_k,
\end{equation}
where $w_k\sim\mathcal{N}(0, Q_k)$ is the process noise.
The measurement model is given by
\begin{equation}
	z_k=\rho_k+v_k,
\end{equation}
where the $v_k\sim\mathcal{N}(0, R_k)$ is the measurement noise.
In each LiDAR scan, the mean $\mu_k$ and variance $\sigma_k$ of the intensities of the ground points are calculated.
The measurement for the innovation computation is then determined as $z_k = \mu_k+2\sigma_k$.

This adaptive approach relies on two assumptions. Firstly, it presumes that nearby consecutive roads should possess similar segmentation coefficients owing to the consistency in ground materials. Secondly, it assumes that the majority of LiDAR points lie on the common asphalt surface, while road marking points exhibit statistically higher intensities.
These two assumptions are satisfied in most urban road environments, ensuring the effectiveness of the approach.
Furthermore, segmenting these high-reflectance points is pivotal for optimizing the efficiency, strategically mitigating the computational load by excluding a significant volume of data points unrelated to road markings.
% This focused segmentation process contributes to a more computationally efficient road marking detection system.

\subsubsection{Probabilistic Local Map Update}

A local map is constructed through the aggregation of spatio-temporally successive LiDAR scans using an odometry, incorporating high-reflectance points to generate a dense point cloud for road marking detection.
However, with the accumulation of scan data, the volume of information grows substantially, leading to an increasing computational burden.

To achieve real-time performance, a novel approach for probabilistically updating the local map is introduced.
This approach employs a probabilistic discarding strategy, wherein each point in the map is selectively removed based on a calculated probability value.
The probability assigned to the $i$-th point in the local map, denoted as $p_i$, is computed by
\begin{equation}
	p_i = \frac{1}{1 + (\left| k - k_i \right| / \eta)^2},
	\label{prob}
\end{equation}
where $k$ denotes the index of the current frame, and $k_i$ represents the frame from which the $i$-th point originates. $\eta$ is a manually-set parameter to determine the probability of discarding old points. As $\eta$ increases, old points are more likely to be retained, resulting in a higher density of points in the probabilistic local map.

As indicated by \eqref{prob}, higher retaining probability values are assigned to newly observed points by the LiDAR sensor.
This strategy effectively ensures the spatio-temporal consistency of the local map, alleviates the impact of accumulated errors over time.
Moreover, when contrasted with the aggregation method employing scans within a fixed window, the proposed approach ensures a more seamless transition in the local map data, thereby yielding higher-quality LiBEV images.

\subsubsection{LiBEV Image Generation}

The generation of the LiBEV image involves dividing the local map into grid cells on the ground plane, where each cell corresponds to a pixel in the LiBEV image.
Within each cell, the RGB value of the corresponding pixel is determined by mapping the maximum intensity value among the enclosed points using a color map.
%Fig.\ref{LiBEV}(a) illustrates an example of the resulting LiBEV image.

Our implementation leverages a proficient instance segmentation network, specifically the CenterMask \cite{centermask}, to accurately segment semantic road markings from the generated LiBEV images.
Subsequently, points located within the grid cells corresponding to the segmented pixels are extracted from the local map.
The extraction yields a semantic point cloud wherein each point is labeled with a specific road marking category.
Notably, our approach is designed to accommodate the segmentation of up to 9 types of road markings, including dashed lanes, solid lanes, stop lines, texts, arrows, diamond signs, triangle sighs, curbs and crosswalks, as shown in Fig. \ref{marking_vis} (b).
The incorporation of diverse semantic road markings, in contrast to approaches solely focused on lane lines, significantly enhances the robustness of map matching-based pose estimation.
In addition, since annotating image semantic segmentation is faster and more convenient than annotating point clouds, the proposed approach converts point clouds into images, which is more conducive to the deployment in practical applications.

%\begin{figure}
%	\centering
%	\subfigure[]{\includegraphics[trim=28 40 28 40, clip, width=\linewidth]{figs/libev.jpg}}
%  \subfigure[]{\includegraphics[trim=0 10 0 42, clip, width=\linewidth]{figs/semanticpoints.jpg}}
%	\caption{(a) The LiBEV image generated from the probabilistic local map. (b) Semantic road markings extracted from the LiBEV using CenterMask.}
%	\label{LiBEV}
%\end{figure}

\subsection{SG-ICP-based Road Marking Registration with HD Map}

After road marking detection, the detected road markings can be associated with their corresponding elements in the HD map shared with the same semantic label. Finally, road marking registration is employed to estimate the pose of the vehicle in 2D space.
In this subsection, the SG-ICP algorithm is introduced for robust registration of detected road markings from LiDAR scans with semantic elements in the HD map.
In our proposed SG-ICP, detected road markings are divided three categories, including lines, line segments and others. Solid lanes and curbs exhibit a linear distribution in their point clouds and lack distinct endpoints, and thus they are classified as lines. Dashed lanes, sidewalks, and stop lines also have a linear distribution but possess endpoints, leading to their classification as line segments. Texts, arrows, diamond signs and triangle signs do not have linear point cloud distribution, and thus they are classified as others.

For lines, the lack of endpoints leads to the complete loss of constraints along the linear direction of these markings. For line segments, endpoints can provide constraints along the linear direction. However, due to inaccurate endpoint estimation, registration between endpoints still leads to significant localization errors along the direction of the line segment.
Consequently, for linear markings, constraints along their linear direction need to have minimal influence on the pose estimation, mitigating the effect of under-constraint issues in the overall pose estimation process.
As for others, their point clouds are not linearly distributed, thus often providing sufficient constraints on the pose estimation. In our algorithm, the registration of the three different categories of markings is organized into a unified representation using the objective function of generalized ICP (GICP).
% Moreover, given the emphasis on detecting road markings on ground surfaces, the subsequent pose estimation is addressed with 3 degrees of freedom (DoF) within a 2D space on the ground plane.

The GICP algorithm incorporates a probabilistic model into the optimization procedure, as defined by
\begin{equation}
	\begin{aligned}
		\boldsymbol{T}^*=
		&\arg\min_{\boldsymbol{T}}
		\bigg(
		\sum_{i=1}^{n}
		(\boldsymbol{q}_{mi}-\boldsymbol{T}\cdot\boldsymbol{q}_{Li})^T\\
		&(\boldsymbol{C}_{mi}+\boldsymbol{R}\boldsymbol{C}_{Li}\boldsymbol{R})^{-1}
		(\boldsymbol{q}_{mi}-\boldsymbol{T}\cdot\boldsymbol{q}_{Li})
		\bigg),
	\end{aligned}
	\label{obj_func}
\end{equation}
where $\boldsymbol{q}_{mi}$ and $\boldsymbol{q}_{Li}$ represent a pair of corresponding points, belonging respectively to the HD map element and the labeled point cloud. Their correspondences are established through the nearest neighbor search strategy in the ICP algorithm. $\boldsymbol{C}_{mi}$ and $\boldsymbol{C}_{Li}$ represent the covariance matrices of points from map and labeled point cloud, respectively, which are appropriately constructed in our semantic GICP (SG-ICP) to mitigate the influence of under-constrained direction.

In our SG-ICP, the probabilistic model is specifically designed by exploiting the semantic and geometric attributes inherent in semantic road markings.
For the points lying on the $i$-th detected road marking instance, the covariance matrix $\tilde{\boldsymbol{C}}_{Li}$ is estimated by
\begin{equation}
	\tilde{\boldsymbol{C}}_{Li} = \frac{1}{n_i - 1}\sum_j^{n_i} (\boldsymbol{p}_{L(i, j)} - \tilde{\boldsymbol{p}}_{Li})\cdot (\boldsymbol{p}_{L(i, j)} - \tilde{\boldsymbol{p}_{Li}})^T,
\end{equation}
where $\boldsymbol{p_{L(i, j)}}$ represents the $j$-th point of the $i$-th road marking instance, $\boldsymbol{\tilde{p_{Li}}}$ represents the centroid of the points. Then, the singular value decomposition (SVD) is performed on $\boldsymbol{C}_{Li}$.
\begin{equation}
\begin{aligned}
	\tilde{\boldsymbol{C}}_{Li}=\boldsymbol{U}_i\tilde{\boldsymbol{\Sigma}}_i\boldsymbol{V}_i,
	\quad
	\tilde{\boldsymbol{\Sigma}}_i=
	\begin{bmatrix}
		\sigma_1^2 & 0 \\
		0 & \sigma_2^2
	\end{bmatrix},
\end{aligned}
\end{equation}
$\sigma_1$ and $\sigma_2$ satisfy $\sigma_1\geq\sigma_2$.
Then, a matrix $\boldsymbol{\Sigma}_i={\rm diag}(1,\epsilon)$ is constructed,
% as
%\begin{equation}
%	\boldsymbol{\Sigma}_i=
%	\begin{bmatrix}
%		1 & 0 \\
%		0 & \epsilon
%	\end{bmatrix},
%\end{equation}
with $\epsilon$ satisfying
\begin{equation}
	\epsilon=
	\begin{cases}
		1e-6, & \text{if the marking is classified lines;}\\
		1e-1, & \text{if the marking is classified line segments;}\\
		1,    & \text{if the marking is classified others.}
	\end{cases}
\end{equation}
The three categories of road markings have distinct values of $\epsilon$, representing the different constraints along the line direction. A value of $\epsilon$ closer to $1.0$ indicates a stronger constraint along the line direction. The final covariance matrix corresponding to the $i$-th road marking can be calculated by
\begin{equation}
	\boldsymbol{C}_{Li}=\boldsymbol{U}_i\boldsymbol{\Sigma}_i\boldsymbol{V}_i.
	\label{CLi}
\end{equation}

The $i$-th semantic element in the HD map is represented as $\{\boldsymbol{v}_{mi}, l_{mi}, P_{mi}\}$, where $\boldsymbol{v}_{mi}$, $l_{mi}$ and $P_{mi} = \{\boldsymbol{p}_{m(i, j)}, j=1, 2, \cdots, n_{mi}\}$ denote the main direction, the semantic label and the point set of the map element, respectively.
The rotation that rotates the basis vector $\boldsymbol{e}_1=[1,0]^T$ to the direction $\boldsymbol{v}_{mi}$ can be calculated by
\begin{equation}
	\boldsymbol{R}_{vi}=\cos\theta\cdot\boldsymbol{I}
	+(1-\cos\theta)\boldsymbol{r}\boldsymbol{r}^T
	+\sin\theta\cdot[\boldsymbol{r}]_\times,
\end{equation}
where
\begin{equation}
	\begin{aligned}
		\boldsymbol{r}=[\boldsymbol{e}_1]_\times\cdot\boldsymbol{v}_{mi},
		\quad
		\theta=\arccos(\boldsymbol{e}_1^T\boldsymbol{v}_{mi}).
	\end{aligned}
\end{equation}
The symbol $[\boldsymbol{r}]_\times$ denote the skew-symmetric matrix associated with the vector $\boldsymbol{r}$.
The covariance matrix corresponding to the $i$-th semantic element is calculated by
\begin{equation}
	\boldsymbol{C}_{mi}=\boldsymbol{R}_{vi}\boldsymbol{\Sigma}_{i}\boldsymbol{R}_{vi}.
	\label{Cmi}
\end{equation}

Finally, associations can be established between the semantic point cloud and the closest points of the map elements shared the same semantic label. Meanwhile, their corresponding covariance matrices calculated in \eqref{CLi} and \eqref{Cmi} are then substituted into the objective function \eqref{obj_func} to initiate the optimization and iteration process. The probabilistic model from SG-ICP characterizes both the semantic and geometric attributes for road marking registration, which improves the accuracy of pose estimation.

\vspace{-5pt}

\section{Experimental Evaluation}

In this section, extensive experiments are conducted using data collected from diverse scenarios and vehicular platforms, demonstrating the accuracy and robustness of the proposed approach across different scenes and types of LiDAR sensors.

\vspace{-7pt}

\subsection{Experimental Setup}

\begin{figure}[htbp]
	\centering
	\subfigure[]{
		\includegraphics[width=0.46\linewidth]{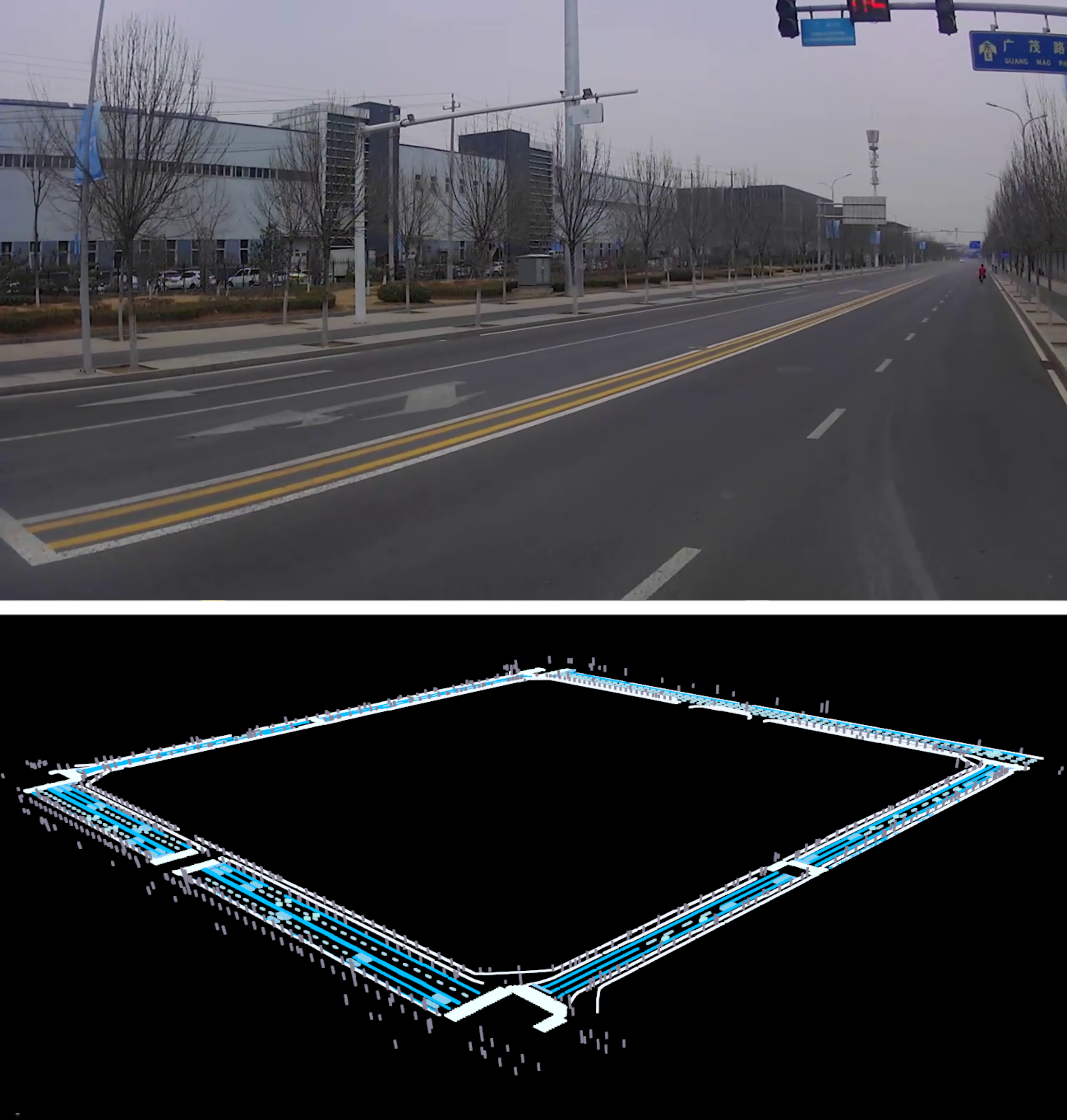}
		\label{fig:subfig1}
	}
	\subfigure[]{
		\includegraphics[width=0.46\linewidth]{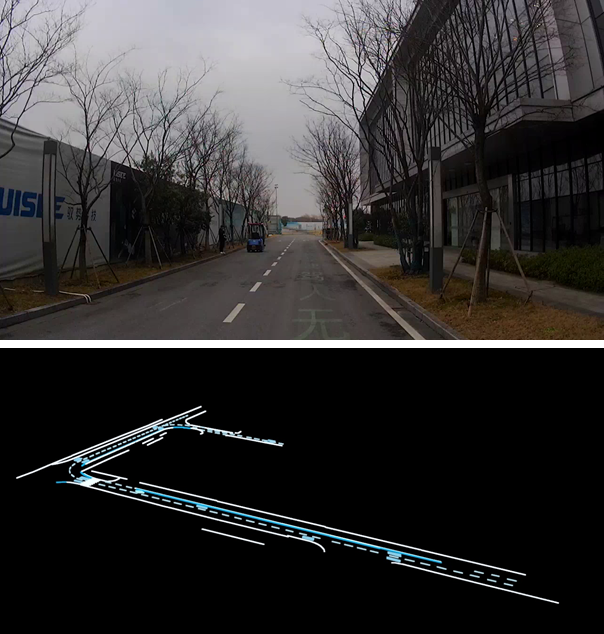}
		\label{fig:subfig2}
	}
	\subfigure[]{
		\includegraphics[width=0.46\linewidth]{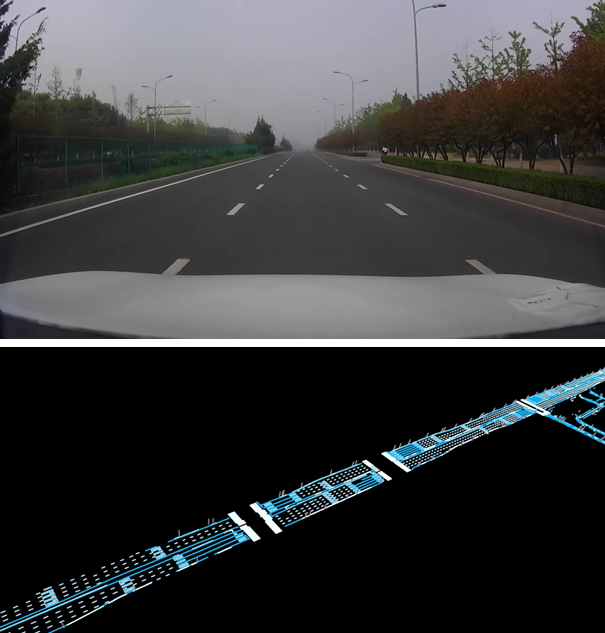}
		\label{fig:subfig3}
	}
	\subfigure[]{
		\includegraphics[width=0.46\linewidth]{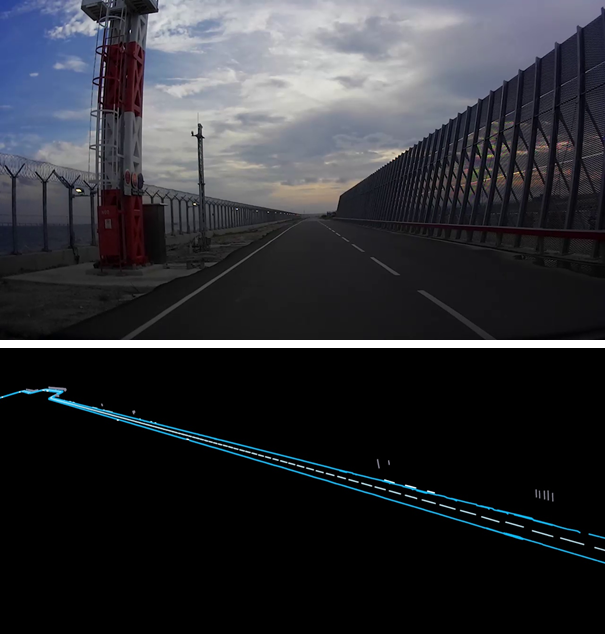}
		\label{fig:subfig4}
	}
	\caption{The experimental scenarios (top) and their corresponding HD maps (bottom).
		(a) \emph{Fangshan1} (b) \emph{Jiashan} (c) \emph{Fangshan2} (d) \emph{Airport}.}
	\label{exp}
\end{figure}

All experiments are conducted on the NVIDIA Jetson AGX Xavier.
The acquisition frequency of LiDAR data is set to $10 Hz$.
The global localization results of vehicles are recorded using Real-Time Kinematic (RTK) and temporally synchronized with the LiDAR data. These RTK results are used as ground truths.
The experimental scenarios and the corresponding HD maps are shown in Fig. \ref{exp}.
\emph{Fangshan1} and \emph{Fangshan2} represent two open urban scenarios in Beijing Fangshan, which covers a $0.30km\times0.25km$ area and spans a length of $2.0km$, respectively.
\emph{Jiashan} depicts an internal road measuring $0.20km$ located in a test field in Zhejiang Jiashan.
\emph{Airport} represents an internal road spanning a length of $4.0km$ located within an airport. For the parameters of our approach, the initial variances of the state-transition model and the measurement model were experimentally set to $0.1$ and $2.0$ in the Kalman filter, respectively. $\eta$ to determine the discarding probability of local map points was set to $50.0$ empirically.

\vspace{-7pt}

\subsection{Evaluation on Road Marking Detection}

\renewcommand\arraystretch{1.2}
\begin{table*}[ht]\small
	\caption{The precision, recall and F1-score for all types of road marking supported by our approach.}
	\begin{center}
		\begin{tabularx}{\textwidth}{m{0.0675\linewidth}<{\centering}
				m{0.0675\linewidth}<{\centering}
				m{0.0675\linewidth}<{\centering}
				m{0.0675\linewidth}<{\centering}
				m{0.0675\linewidth}<{\centering}
				m{0.0675\linewidth}<{\centering}
				m{0.0675\linewidth}<{\centering}
				m{0.0675\linewidth}<{\centering}
				m{0.0675\linewidth}<{\centering}
				m{0.0675\linewidth}<{\centering}
				m{0.0675\linewidth}<{\centering}}
			\toprule
			& Dashed lane & Solid lane & Stop line & Text & Arrow & Diamond sign & Triangle sign & Curb & Crosswalk & Average \\
			\hline
			Precision & 94.05\% & 86.11\% & 81.16\% & 89.19\%  & 91.70\% & 94.66\% & 88.23\%  & 64.52\% & 72.77\% & 86.31\% \\
			Recall    & 96.50\% & 85.48\% & 75.47\% & 100.00\% & 96.57\% & 97.37\% & 100.00\% & 76.11\% & 71.96\% & 88.56\% \\
			F1-score  & 95.27\% & 85.79\% & 78.21\% & 94.29\%  & 94.07\% & 95.99\% & 93.75\%  & 69.83\% & 72.36\% & 85.20\% \\
			\bottomrule
		\end{tabularx}
		\label{marking}
	\end{center}
\end{table*}

In this subsection, an experiment is conducted to assess the performance of our road marking detection approach using precision-recall metrics.
To ensure a comprehensive evaluation, 80\% of the manually annotated LiBEV data is randomly selected for training, while the remaining 20\% is reserved for testing.
The manual annotations serve as the ground truths against which we evaluate the precision and recall of our approach in detecting road markings.
A true positive sample is identified when the Intersection over Union (IoU) between the detected instance and its corresponding annotated instance exceeds 0.5, and both instances shared the same semantic label.
Conversely, a false positive sample represent a detection result for which no corresponding instance with the same semantic label and an IoU greater than 0.5 could be found in the ground truths.
Meanwhile, a false negative sample indicates that an instance present in the ground truth is not successfully detected by our approach.

The precision, recall and F1-score for all types of road markings supported by the approach are presented in TABLE \ref{marking}.
The proposed approach successfully detects 9 distinct types of road markings, assigning semantic labels to each point in the LiDAR data, as visually depicted in Fig. \ref{marking_vis} (b).
The experimental results demonstrate the effectiveness of our approach in successfully detecting common road elements, achieving high precision and recall rates.
Notably, certain elements such as curbs and crosswalks exhibit a slight decrease in precision, attributed to their visual similarity to lane markings in LiBEV images.
However, the subsequent HD map registration steps effectively mitigate the impact of these false positives on localization.
Moreover, the proposed detection approach is highly efficient, meeting real-time perception requirements for vehicles, which is detailed in Section \ref{eval_runtime}.

\begin{table*}[htbp]\small
	\caption{Average localization error compared to the ground-truth obtained through RTK.}
	\begin{center}
		\begin{tabular}{cccccccccccc}
			\toprule
			Sequence & Scene & LiDAR type & \multicolumn{2}{c} {Longitudinal error (m)} & \multicolumn{2}{c} {Lateral error (m)} & \multicolumn{2}{c} {Yaw error (deg)} \\

			& & & ICP & SG-ICP & ICP & SG-ICP & ICP & SG-ICP \\
			\hline
			S1 & \emph{Fangshan1} & 2\#Hesai-Pandar64       & 0.158 & \textbf{0.137} & 0.050 & \textbf{0.043} & 0.233 & \textbf{0.208} \\
			S2 & \emph{Fangshan1} & 1\#Hesai-Pandar64       & 0.165 & \textbf{0.160} & 0.051 & \textbf{0.041} & 0.386 & \textbf{0.346} \\
			S3 & \emph{Fangshan1} & 2\#VLP-32C              & 0.167 & \textbf{0.130} & 0.127 & \textbf{0.043} & 0.302 & \textbf{0.188} \\
			S4 & \emph{Fangshan1} & 2\#HAP                  & 0.164 & \textbf{0.139} & 0.080 & \textbf{0.058} & 0.299 & \textbf{0.218} \\
			S5 & \emph{Jiashan}    & 2\#VLP-32C + 1\#VLP-16 & 0.082 & \textbf{0.077} & 0.055 & \textbf{0.050} & 0.547 & \textbf{0.401} \\
			S6 & \emph{Jiashan}    & 3\#HAP                  & \textbf{0.099} & 0.106 & 0.082 & \textbf{0.062} & 0.330 & \textbf{0.309} \\
			S7 & \emph{Fangshan2}    & 1\#Hesai-Pandar64       & \textbf{0.124} & 0.125 & 0.091 & \textbf{0.040} & 0.277 & \textbf{0.184} \\
			S8 & \emph{Airport}    & 2\#Hesai-XT16           & 0.129 & \textbf{0.128} & 0.116 & \textbf{0.050} & 0.447 & \textbf{0.230} \\
			\bottomrule
		\end{tabular}
		\label{quantitaty}
	\end{center}
\end{table*}

\vspace{-7pt}

\subsection{Evaluation on Localization}

The SG-ICP algorithm proposed in this paper is assessed based on lateral, longitudinal, and yaw errors.
The evaluation encompasses eight experimental sequences, spanning four scenarios and employing seven different LiDAR configurations, demonstrating the flexibility of the proposed approach.
The widely-used ICP algorithm is chosen as the baseline for evaluation, and the comparative results are presented in TABLE \ref{quantitaty}.
As indicated in the table, the SG-ICP algorithm outperforms the ICP-based approach in most sequences.
Notably, SG-ICP has a clear superiority in terms of lateral and yaw accuracy, due to the emphasis placed on the sufficiently constrained direction during the SG-ICP calculation process.

\begin{figure}[htbp]
	\centering
	\subfigure[S1]{
		\includegraphics[width=0.44\linewidth]{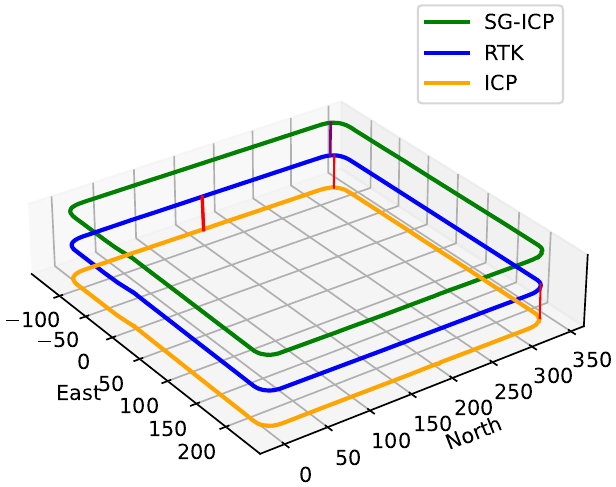}
	}
	\subfigure[S2]{
		\includegraphics[width=0.42\linewidth]{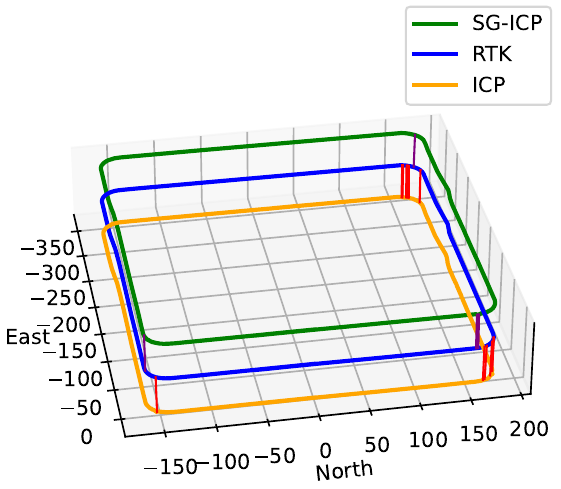}
	}
	\subfigure[S3]{
		\includegraphics[width=0.42\linewidth]{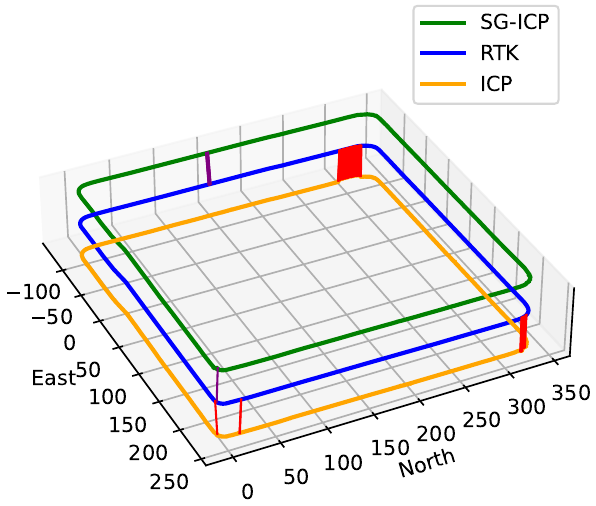}
	}
	\subfigure[S4]{
		\includegraphics[width=0.42\linewidth]{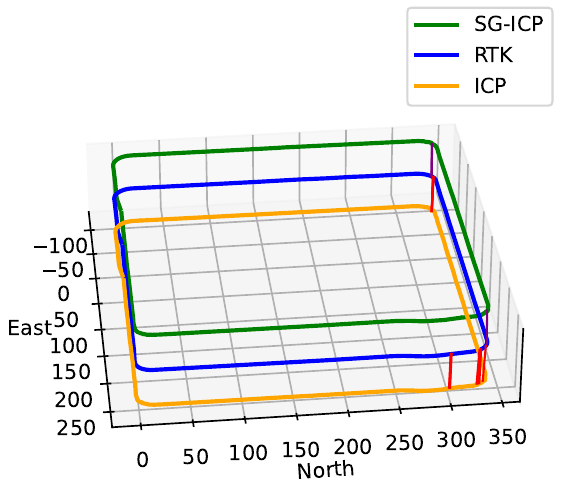}
	}
	\subfigure[S5]{
		\includegraphics[width=0.42\linewidth]{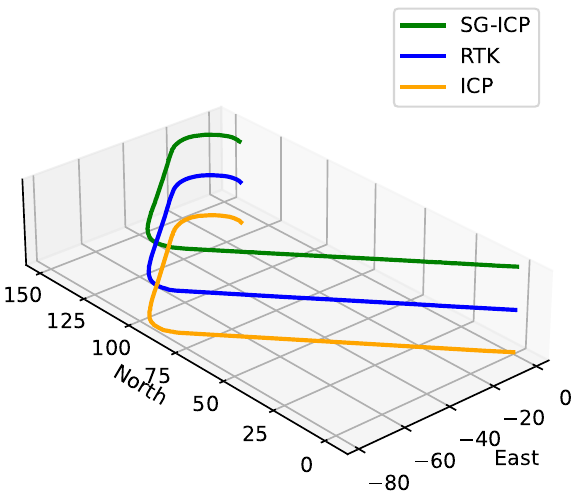}
	}
	\subfigure[S6]{
		\includegraphics[width=0.36\linewidth]{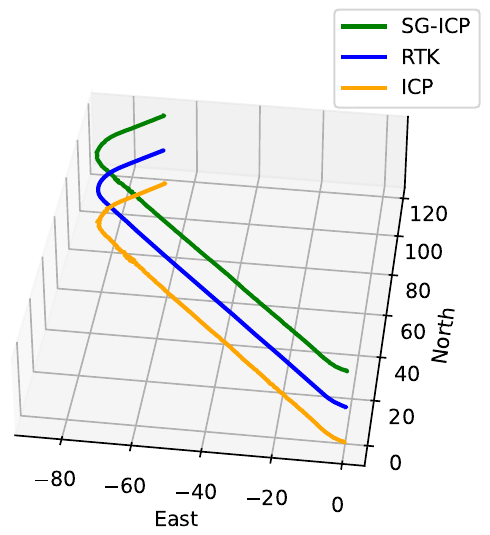}
	}
	\subfigure[S7]{
		\includegraphics[width=0.42\linewidth]{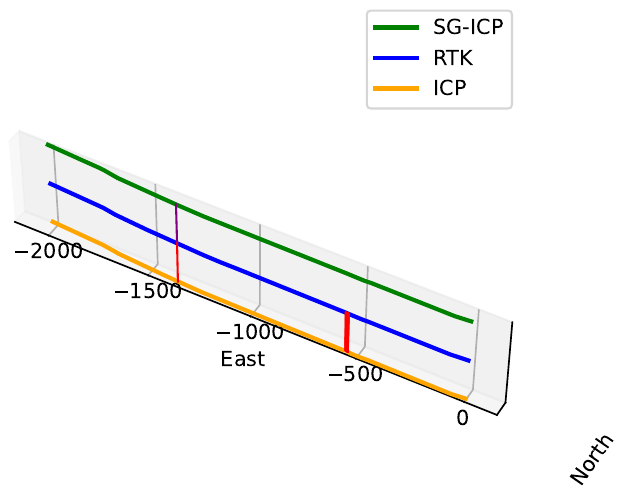}
	}
	\subfigure[S8]{
		\includegraphics[width=0.42\linewidth]{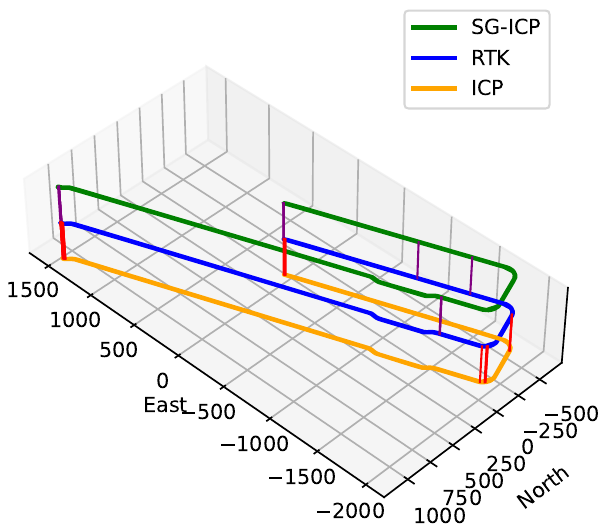}
	}
	\caption{Comparison between the trajectories estimated by SG-ICP and ICP is conducted using ground-truth trajectories provided by RTK. The substantial localization error of SG-ICP and ICP are marked with purple and red lines, respectively, where estimated distance errors exceed 2.0 m or yaw errors surpass 5.0\degree.}
	\label{visualization}
\end{figure}

Fig. \ref{visualization} depicts the visualized trajectories estimated by SG-ICP-based and ICP-based approaches, respectively, in comparison with the ground-truths acquired through RTK.
It is worth noting that the substantial localization error of SG-ICP and ICP are marked with purple and red lines, respectively, where estimated distance errors exceed 2.0 m or yaw errors surpass 5.0\degree.
It is evident from Fig. \ref{visualization} that SG-ICP demonstrates significantly fewer occurrences of substantial localization errors compared to ICP across all sequences.

In conclusion, the proposed approach achieves centimeter-level lateral localization accuracy in a variety of environmental scenarios and with different types of LiDAR sensors.
The tested sensors encompass not only traditional mechanical LiDARs like VLP-32C, Hesai-Pandar64, and Hesai-XT16 but also solid-state LiDARs such as HAP.
The comprehensive experiments illustrate the robustness and adaptability of the approach across diverse scenarios and sensor types.
In addition, it is worth noting, as indicated in TABLE \ref{quantitaty}, that the longitudinal error is slightly larger than the lateral error.
In the urban road scenario where autonomous driving occurs, the majority of road markings exhibit a linear shape along the longitudinal direction.
Consequently, the stronger influence of lateral constraints, compared to longitudinal constraints, contributes to a more accurate and precise lateral localization outcome.
Nevertheless, our approach ensures that the longitudinal error remains below 0.20 m, thereby ensuring its effectiveness in autonomous driving applications.

\vspace{-7pt}

\subsection{Evaluation on Runtime}
\label{eval_runtime}

\begin{figure}[htbp]
	\centering
	\includegraphics[width=0.85\linewidth]{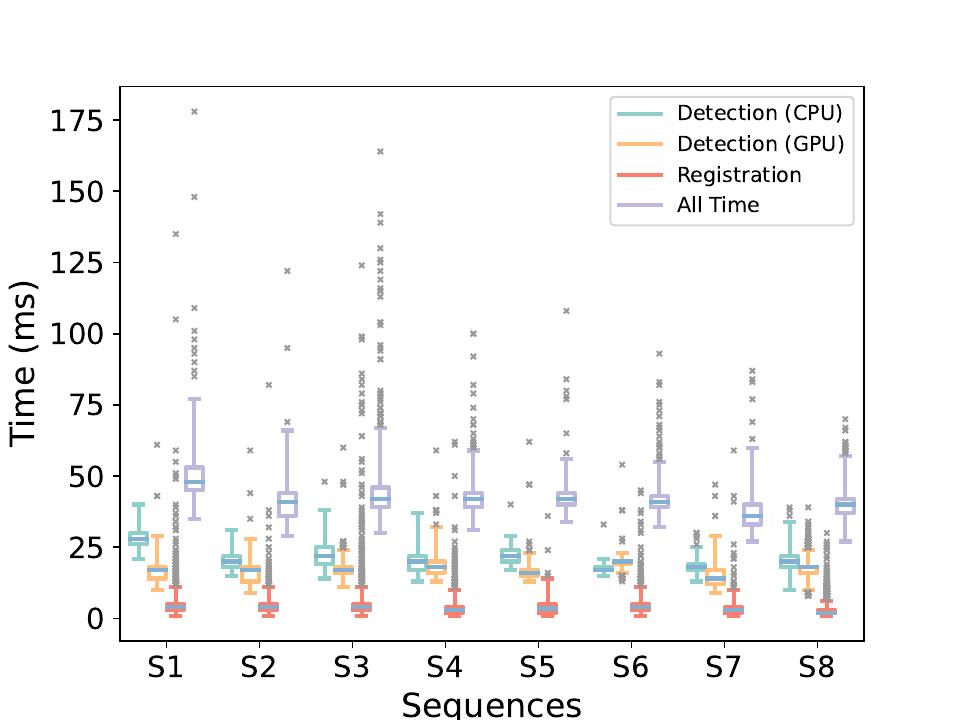}
	\caption{A box plot illustrating the time consumption for each sub-step of the proposed approach.}
	\label{boxplot}
\end{figure}

During the experiments conducted on the eight sequences, the runtime for each sub-step of our approach is detailed in TABLE \ref{time}.
The corresponding box-plot depicting the statistical results can be observed in Fig. \ref{boxplot}. It is worth noting that the runtime of the detection sub-step is divided into CPU time and GPU time. CPU time refers to the time consumed by the steps processed by the CPU, including high-reflectance point segmentation, probabilistic local map update, and LiBEV image generation. GPU time refers to the inference time of the instance segmentation of the LiBEV image. The registration sub-step is processed only by the CPU.
It can be seen that, when utilizing the onboard processor XAVIER, the average and maximum runtime of the overall approach consistently remains below 50 ms and 200 ms across various scenes and types of LiDAR sensors.
% This performance level surpasses the data acquisition frequency of the LiDAR sensor.
Consequently, the efficiency of the proposed system proves sufficient for real-time perception and localization in autonomous vehicle applications.

\renewcommand\arraystretch{1.2}
\begin{table}[htbp]\small
	\caption{The runtime for each sub-step of the proposed approach.}
	\begin{center}
		\begin{tabular}{m{0.08\linewidth}<{\centering}m{0.17\linewidth}<{\centering}m{0.17\linewidth}<{\centering}m{0.17\linewidth}<{\centering}m{0.17\linewidth}<{\centering}}
			\toprule
			Seq. & Detection (CPU) & Detection (GPU) & Registration &  Total Time Cost \\
			\hline
			S1 & 28.20ms & 16.47ms & 4.63ms & 49.30ms \\
			S2 & 20.34ms & 16.36ms & 4.25ms & 40.95ms \\
			S3 & 22.13ms & 16.42ms & 4.43ms & 42.98ms \\
			S4 & 19.66ms & 18.21ms & 3.48ms & 41.35ms \\
			S5 & 22.20ms & 16.55ms & 3.96ms & 42.71ms \\
			S6 & 17.46ms & 19.13ms & 4.73ms & 41.32ms \\
			S7 & 18.69ms & 14.74ms & 3.46ms & 36.89ms \\
			S8 & 19.98ms & 16.75ms & 2.63ms & 39.36ms \\
			\bottomrule
		\end{tabular}
		\label{time}
	\end{center}
\end{table}

Moreover, it is worth highlighting that the runtime on the \emph{S1} sequence is only 8.35 ms longer than that on the \emph{S2} sequence, despite the fact that the data quantity of \emph{S1} is twice that of \emph{S2} (as indicated in the \emph{LiDAR type} column in TABLE \ref{quantitaty}).
This observation demonstrates that the runtime does not exhibit a linear increase with the quantity of the point cloud data, because the substantial reduction in the quantity of the aggregated local map points is achieved through a probabilistic discarding strategy.
% This strategy involves removing points with less information based on an adaptively calculated probability value.
% Consequently, the system is able to maintain real-time performance even as the data quantity increases.

\vspace{-7pt}

\subsection{Evaluation on Robustness}

To demonstrate the robustness of our approach, we evaluate the localization errors at different vehicle speeds, as outlined in TABLE \ref{speed}.
In particular, the vehicle was driven at speeds of 20 km/h, 40 km/h, and 60 km/h in the \emph{Fangshan1} scenario using 1 Hesai-Pandar64 LiDAR.
The obtained results were then compared against the ground-truth provided by RTK.
As evident from TABLE \ref{speed}, there is a slight increase in the localization error with higher driving speeds.
This can be attributed to the fact that, as the driving speed increases, the point cloud data captured by the LiDAR sensors is more prone to motion distortions.
Despite the slight increase in localization error with higher driving speeds, the proposed approach consistently maintains a relatively high level of localization accuracy.
This demonstrates the robustness of the approach across varying vehicle speeds.
Regarding real-time performance, as indicated in TABLE \ref{speed}, the overall system runtime is minimally affected by increases in driving speed.
This further highlights the robustness of the system in handling variations in vehicle speed.

\renewcommand\arraystretch{1.2}
\begin{table}[htbp]\small
	\caption{The localization errors at various driving speeds.}
	\begin{center}
		\begin{tabular}{m{0.13\linewidth}<{\centering}m{0.17\linewidth}<{\centering}m{0.14\linewidth}<{\centering}m{0.15\linewidth}<{\centering}m{0.15\linewidth}<{\centering}}
			\toprule
			Speeds & {Longitudinal error (m)} & {Lateral error (m)} & {Yaw error (deg)} & {Time cost (ms)} \\
			\hline
			20 km/h & 0.166 & 0.048 & 0.366 & 42.56 \\
			40 km/h & 0.124 & 0.067 & 0.620 & 44.40 \\
			60 km/h & 0.153 & 0.091 & 0.775 & 44.92 \\
			\bottomrule
		\end{tabular}
		\label{speed}
	\end{center}
\end{table}

\begin{figure}[htbp]
	\centering
	\subfigure[]{
		\includegraphics[width=0.46\linewidth]{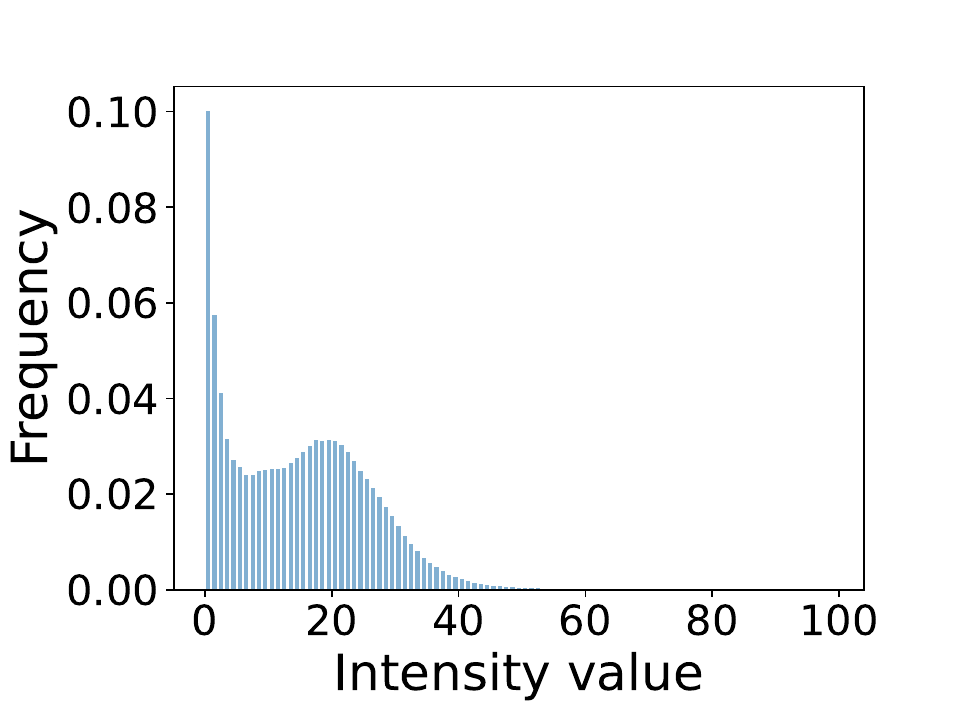}
	}
	\subfigure[]{
		\includegraphics[width=0.46\linewidth]{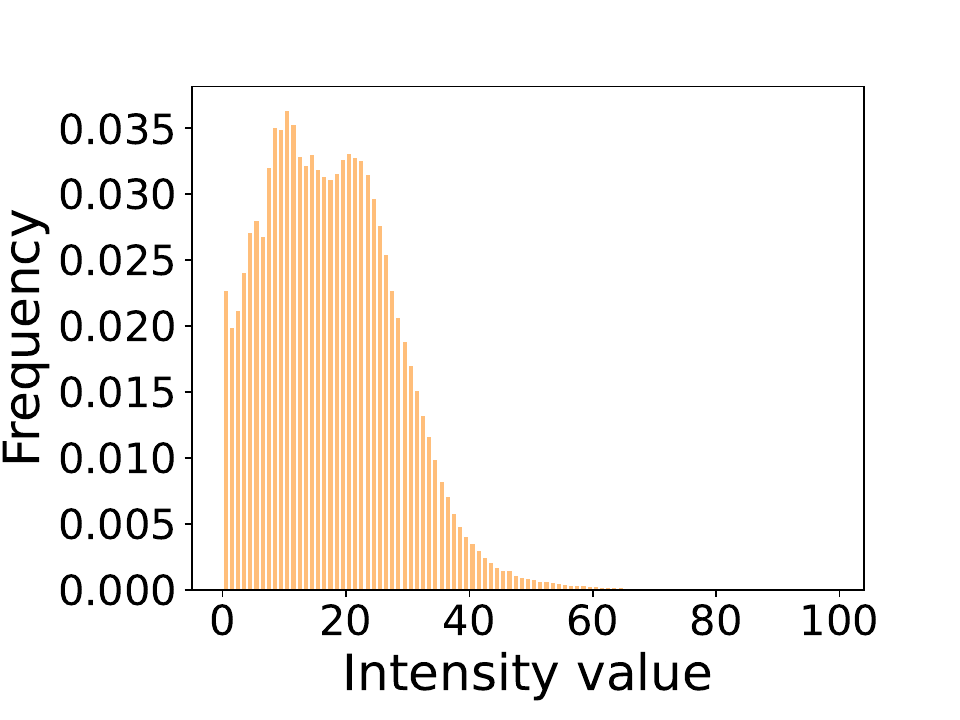}
	}
	\caption{Histograms illustrating the LiDAR intensity distribution on (a) sunny and (b) rainy days, respectively, in the \emph{Fangshan1} scenario.}
	\label{weather_inst}
\end{figure}

\begin{figure}[htbp]
	\centering
	\subfigure[]{
		\includegraphics[width=0.46\linewidth]{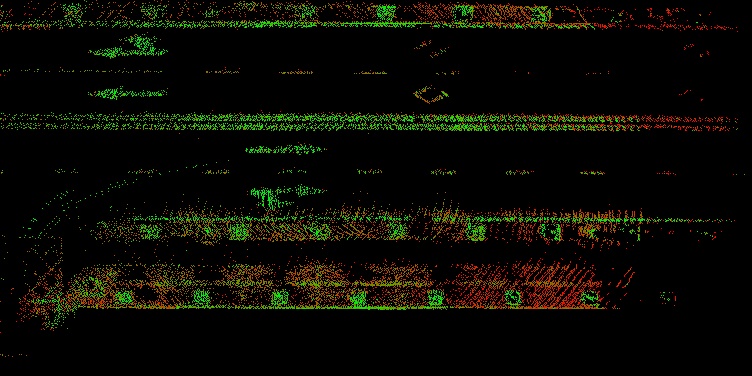}
	}
	\subfigure[]{
		\includegraphics[width=0.46\linewidth]{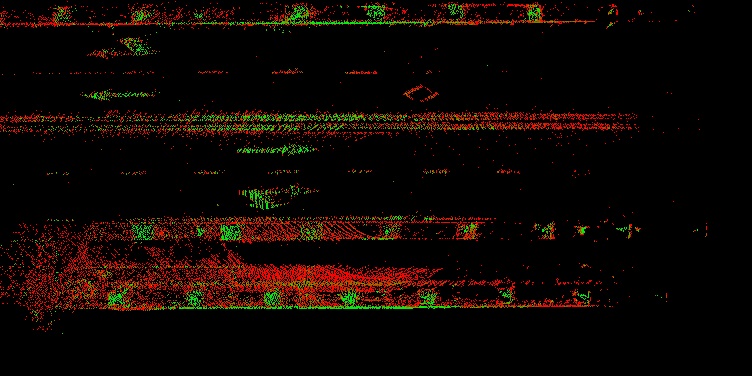}
	}
	\caption{The LiBEV images on (a) sunny and (b) rainy days in the \emph{Fangshan1} scenario.
			Pixel coloring is based on intensity values,
			with green representing high intensity and red representing low intensity.}
	\label{weather_bev}
\end{figure}

To illustrate the robustness of our approach under varying weather conditions, experiments were conducted in different settings.
As depicted in Fig. \ref{weather_inst}, the intensity distribution of LiDAR point clouds on dry and wet road surfaces (on sunny and rainy days) typically exhibits significant differences.
As a result, rainy weather poses considerable challenges to intensity-based road marking extraction, particularly for methods relying on fixed intensity thresholds.
The LiBEV images generated under both sunny and rainy weather conditions are depicted in Fig. \ref{weather_bev}.
% Pixel coloring is based on intensity values, with green representing a relatively high intensity and red representing low intensity.
It is evident that the proposed adaptive threshold-based approach consistently provides stable and accurate segmentation results, even in the presence of significantly different intensity distributions caused by varying weather conditions.
TABLE \ref{weather_tab} presents a comparison of localization errors under both dry and wet ground conditions in the \emph{Fangshan1} scenario, employing 1 Hesai-Pandar64 LiDAR. Although more noise in LiBEV images causes an increase in localization error when driving on wet ground, it can still ensure average lateral error within 0.10 m and longitudinal error within 0.20 m.
These results demonstrate the robustness of our approach in addressing challenging weather conditions.

\renewcommand\arraystretch{1.2}
\begin{table}[htbp]\small
	\caption{The localization errors at various road conditions.}
	\begin{center}
		\begin{tabular}{m{0.15\linewidth}<{\centering}m{0.2\linewidth}<{\centering}m{0.15\linewidth}<{\centering}m{0.2\linewidth}<{\centering}}
			\toprule
			Speeds & {Longitudinal error (m)} & {Lateral error (m)} & {Yaw error (deg)} \\
			\hline
			Dry road & {0.160} & {0.041} & {0.346} \\
			Wet road & 0.199 & 0.056 & 0.390 \\
			\bottomrule
		\end{tabular}
		\label{weather_tab}
	\end{center}
\end{table}

\section{Conclusion}

In this paper, we introduce a LiDAR-based online environmental perception and localization system with high efficiency and robustness.
The proposed road marking detection approach employs a novel adaptive segmentation technique to enhance efficiency, and utilize a spatio-temporal probabilistic local map to ensure the density of points.
For road marking registration, an SG-ICP algorithm is designed, modeling linear road markings as 1-manifolds embedded in 2D space.
Our approach minimizes the influence of constraints along the linear direction of markings, to address the under-constrained problem, and thus improve the localization accuracy.
Extensive experiments conducted in real-world urban environments demonstrate the effectiveness and robustness of the proposed system, showcasing its potential for reliable online environmental perception and localization.
However, our approach cannot be applied to roads without road markings on the ground surface, due to the lack of high-reflectance points. In future work, we will explore the effective utilization of above-ground information to improve the robustness of localization.

\bibliographystyle{IEEEtran}
\bibliography{bare_jrnl_new_sample4}

\end{document}